\documentclass[sigconf,natbib=false]{acmart}

\AtBeginDocument{%
  }

\copyrightyear{2024}

\RequirePackage[
  datamodel=acmdatamodel,
  style=acmnumeric,
  ]{biblatex}

\begin{document}

\title{The Solution for The PST-KDD-2024 OAG-Challenge}

\author{Shupeng Zhong}
\affiliation{%
  \institution{Nanjing University of Science and Technology}
  \city{Nanjing Shi}
  \country{China}}
\email{zspnjlgdx@njust.edu.cn}

\author{Xinger Li}
\affiliation{%
  \institution{Nanjing University of Science and Technology}
  \city{Nanjing Shi}
  \country{China}
}

\author{Shushan Jin}
\affiliation{%
  \institution{Nanjing University of Science and Technology}
  \city{Nanjing Shi}
  \country{China}
}

\author{Yang Yang}
\affiliation{%
  \institution{Nanjing University of Science and Technology}
  \city{Nanjing Shi}
  \country{China}}
\email{yyang@njust.edu.cn}

\begin{abstract}
\quad In this paper, we introduce the second-place solution in the KDD-2024 OAG-Challenge paper source tracing track. Our solution is mainly based on two methods, BERT and GCN, and combines the reasoning results of BERT and GCN in the final submission to achieve complementary performance. In the BERT solution, we focus on processing the fragments that appear in the references of the paper, and use a variety of operations to reduce the redundant interference in the fragments, so that the information received by BERT is more refined. In the GCN solution, we map information such as paper fragments, abstracts, and titles to a high-dimensional semantic space through an embedding model, and try to build edges between titles, abstracts, and fragments to integrate contextual relationships for judgment. In the end, our solution achieved a remarkable score of 0.47691 in the competition.
\end{abstract}

\maketitle

\section{Introduction}
\quad With the rapid development of science and technology, the number of papers has exploded. For academic papers in various fields, it is very difficult to find the technical development process from a large amount of literature. The source paper tracing task aims to find important references in the paper, that is, whether they have made inspiring contributions to the ideas or main methods of the paper. At present, the solutions to this task are mainly divided into three categories: large models, graph convolutional neural networks (GCNs), and machine learning. Among them, thanks to the stronger robustness and generalization of large models~\cite{yang2023contextualized,yang2021corporate}, and the powerful relationship modeling and feature aggregation capabilities of graph convolutional neural networks~\cite{liu2023qtiah,yang2018multi}, their excellent performance has attracted more attention from researchers.

Under this background, the KDD-2024 OAG-Challenge PST paper source tracing task requires contestants to design an efficient and accurate algorithm to identify the source paper \cite{zhang2024pst,zhang2024oag}. Here we will introduce our second place solution.

Our main contributions are as follows:

\begin{itemize}

\item For extremely redundant XML format context data, we designed data cleaning methods such as removing XML characters with unclear semantic information, embedding text content prompts, and setting dynamic context acquisition length, so that the context used to train BERT-like model will contain richer classification semantic information.

\item In terms of GCN, we constructed a relationship network between reference title information, paper title information, paper abstract information, and paper context information, and performed node classification with the help of GCN's powerful relationship modeling capabilities.

\end{itemize}

Finally, we integrated the reasoning results of BERT and GCN to achieve complementary performance. We believe that the combination of BERT's powerful context understanding ability and GCN's ability to integrate full-text structural information has greatly improved classification accuracy.



\section{Related Work}
\quad Understanding scientific trends and the flow of ideas is critical to both policy development by funding agencies and knowledge discovery by researchers. Citation relationship analysis reveals the trajectory of scientific evolution, but there is a significant gap between large-scale semantically rich citation relationships and the backbone structure of scientific evolution. To reveal the context in which science develops, we need to simplify citation graphs to track the sources of publications and reveal heuristic relationships between papers.

Tang et al~\cite{Tang}. studied citation semantics, defined three categories of citation links, and constructed a computer science dataset containing approximately 1000 citation pairs. Valenzuela et al~\cite{Valenzuela}. present a new dataset of 450 citation pairs and distinguish between incidental and important citations. Jurgens et al~\cite{Jurgens}. proposed a larger dataset in the NLP field that contained nearly 2000 citation pairs, but less than 100 citation pairs were annotated as motivations. 

Pride and Knoth believe that abstract similarity is one of the most indicative features for predicting citation importance. Hassan et al~\cite{Hassan}. used random forest to evaluate the importance of citations and combined context-based and clue word-based features. He et al~\cite{He}. adapted the LDA model to the citation network and developed a new inherited topic model to describe topic evolution. With the rapid development of large models\cite{yang2019semi, yang2019semi1, yang2018complex, yang2019deep,yang2021corporate}, image and text features are constantly being learned by large models\cite{fu2024noise,wancovlr,yang2021corporate,yang2021learning,wancovlr,yang2021cost,yang2021s2osc}. Färber et al~\cite{Färber}. proposed an approach based on convolutional recurrent neural networks to classify potential citation contexts. Jiang and Chen proposed a SciBERT-based contextual representation model to classify citation intent and achieved over 90\% AUC on some datasets. Yin et al~\cite{Yin}. designed MRT, an unsupervised framework to identify important previous work or evolutionary trajectories by building an annotated evolutionary roadmap. The framework utilizes generative expression embedding and clustering methods to group publications to capture potential relationships between publications. 


\section{Method}
\quad The training dataset released for this task is mainly given in the form of XML papers. According to our observation, XML data contains a lot of redundant information, so extra attention needs to be paid to data cleaning. Inspired by the ~\cite{zhang2024pst}, this article divides the paper tracing task into two aspects: BERT-based text classification task and GCN-based node classification task  .

The BERT-like text classification task performs text classification by understanding the context in which the reference appears based on text cleaning. The GCN node classification task focuses on understanding the relationship between the reference title and the paper title based on the content of the reference fragment and the paper abstract.

\subsection{Text sequence classification task based on BERT-like model}
\quad The text directly regards the paper tracing task as a text sequence binary classification task, that is, the probability that a certain paper is the source paper. This article mainly cleans the XML format training data and does not modify the model module. The official Json file for training contains 788 papers including source papers and reference papers. We only use 2/3 of the papers for training and the remaining 1/3 for verification.

\begin{figure}[h]
  \centering
  \includegraphics[width=\linewidth]{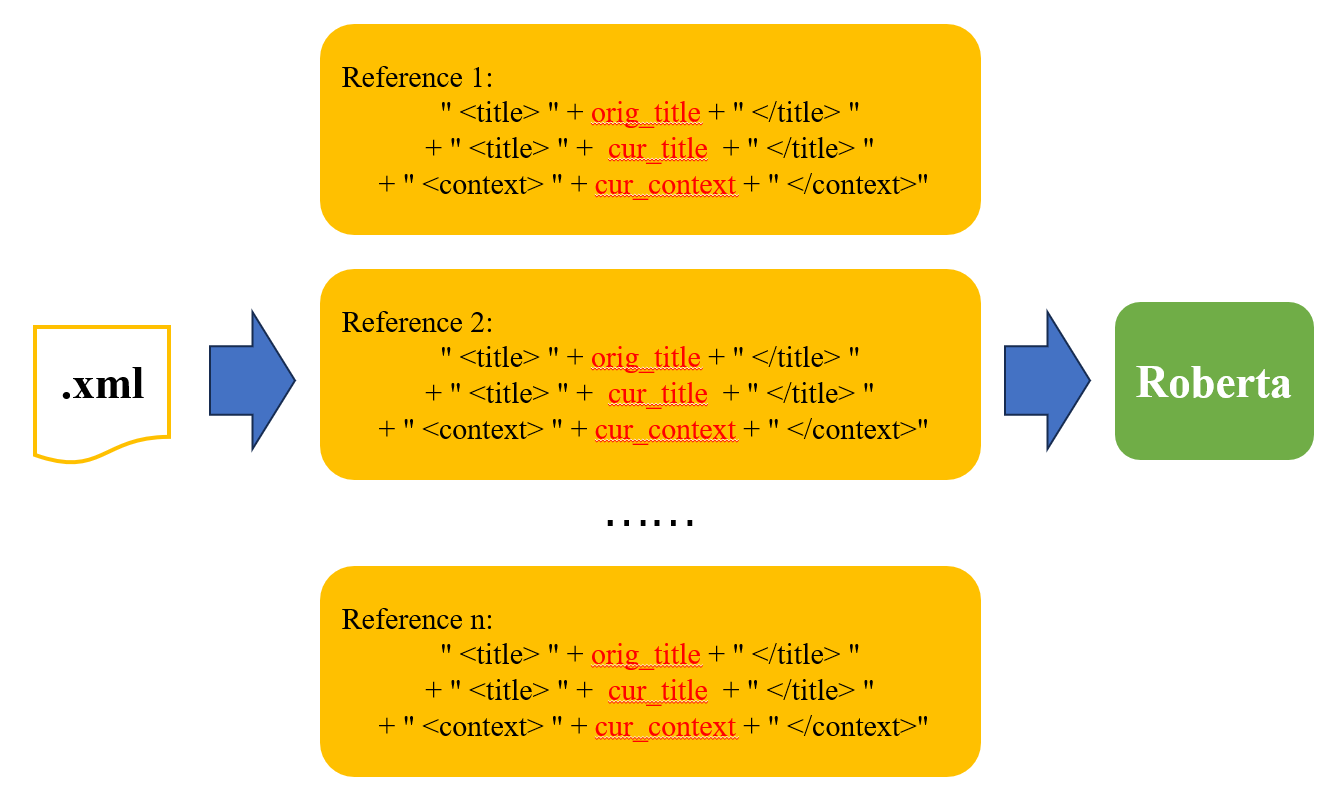}
  \caption{Train Dataset For Roberta}
\end{figure}

In the text cleaning process, we mainly perform the following operations:

\begin{itemize}
\item When finally forming the training data, the original title, the citation title, and the context of the citation in the original text are inserted in the following format, as show in figure 1.

\item The XML format context is cleaned , such as removing line breaks, "<", ">", special characters, and other characters whose semantic information is unclear to humans.

\item Considering that the same citation appears more than once in the original text, we split the multiple citations by spaces and concatenate them into the final citation context, namely the cur\_context part .

\item We have added a hint for the reference section, such as "[Introduction]".

\item We used two text truncation methods, one is to use semantics to truncate the complete sentence, and the other is to use the absolute number of characters before and after. We found that the combination of the two methods can improve the performance of the model .

\item For multiple references appearing in the same segment, we set the references other than the target reference to null.

\end{itemize}

\subsection{GCN-based node classification task}
\quad Graph Convolutional Network (GCN) is a neural network model for processing graph structured data. GCN extracts features on the graph through convolution operations, which can effectively capture the relationship between nodes and their neighbors. Unlike traditional convolutional neural networks, GCN is specially designed for graph data and can process non-Euclidean structured data, such as social networks, knowledge graphs, and citation networks.

The core idea of GCN is to aggregate the features of each node with the features of its neighboring nodes through convolution operations, thereby updating the representation of the node. The forward propagation formula of GCN is as follows:

\begin{equation}
    H^{(l+1)} = \sigma\left( D^{-1/2} A D^{-1/2} H^{(l)} W^{(l)} \right)
\end{equation}

where:
\begin{itemize}
    \item \(H^{(l)}\) denotes the node feature matrix at the \(l\)-th layer.
    \item \(A\) is the adjacency matrix of the graph.
    \item \(D\) is the degree matrix of the adjacency matrix.
    \item \(W^{(l)}\) is the weight matrix at the \(l\)-th layer.
    \item \(\sigma\) is an activation function (such as ReLU).
\end{itemize}

\begin{figure}[h]
  \centering
  \includegraphics[width=\linewidth]{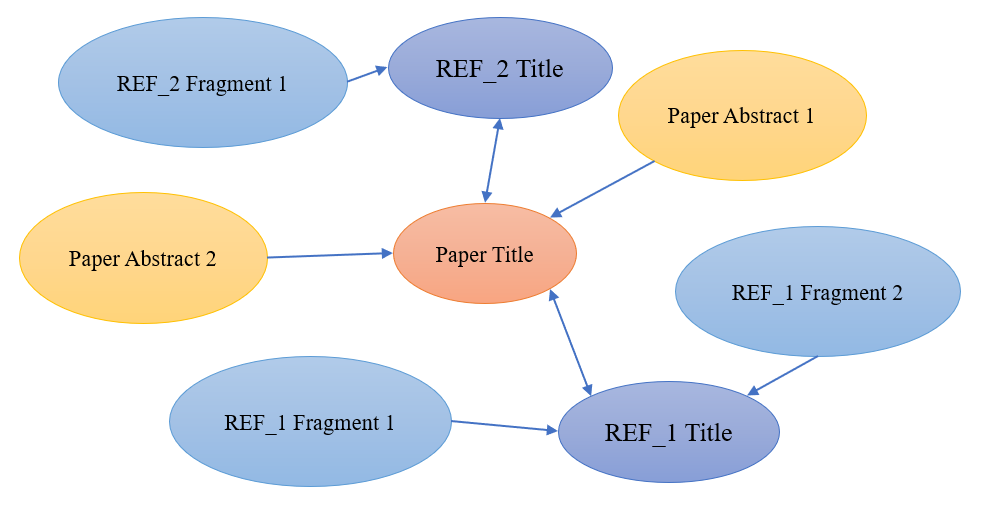}
  \caption{Our GCN Mapping Method}
\end{figure}

Drawing on the research of Yao et al~\cite{Yao1}. and Wu et al~\cite{Kun}. on GCN text classification and combining GCN's powerful relationship modeling and feature aggregation capabilities with its excellent efficiency, we used GCN to classify text in the competition. 

Figure 2 shows how we use GCN to build graphs for papers. We use the Spacy library to semantically segment each paragraph of the paper, then build a unidirectional edge between the paper abstract and the paper title, a bidirectional edge between the reference title and the paper title, and a unidirectional edge between the reference and the text block where the reference appears. The node information is a high-dimensional vector of the text processed by the embedding model. 

As for the direction of edge building, according to our experiments, the effect of using all bidirectional edge building or all unidirectional edge building is reduced to varying degrees.

\section{Experiment}

\quad We conducted experiments on GCN and BERT-like models respectively.

\subsection{BERT-like models}
\quad Here we present our experimental results on BERT-like models.We mainly tried two models: SCIBERT~\cite{beltagy2019scibert,devlin2018bert} and his variants CS\_ROBERTA\_BASE~\cite{liu2019roberta}

\textbf{Dataset} The data is provided by the official. The data includes training Json files and a large number of XML format papers. The Json files contain the basic information of 788 papers such as ID, title, references, and source papers; the XML format papers are converted from the PDF files of the original papers using Grobid. It is worth noting that although the official allows the use of OAG and DBLP Citation data, this article does not use them.

\textbf{Details} Considering that the task background is academic papers in various fields, we tried to use models such as SCIBERT and CS\_ROBERTA\_BASE that have been trained on a large amount of scientific text. Eventually, we found that CS\_ROBERTA\_BASE achieved the best results on the locally partitioned Train, val, official Val, and Test. We only used one NVIDIA RTX 4090 to train the CS\_ROBERTA\_BASE model, setting the initial learning to 5e-6, batch\_size=24, epoch=10 and implemented early stopping after 5 epochs when performance did not continue to improve. Finally, we selected the model with the highest local AP value as the optimal model.

\textbf{Results} Table 1 shows the performance of our method on the official final test set. We add the method step by step according to the order of the labels in the table. The experimental results fully confirm the effectiveness of the method. The content cleaning of the XML format context removes a lot of redundant information. The embedding of the paper title and the corresponding prompt words makes the model pay more attention to and understand the corresponding content. At the same time, the CS\_ROBERTA\_BASE model has larger parameters and stronger learning ability.

\begin{table}
  \caption{Experimental results of different methods on BERT-like models}
  \label{tab:freq}
  \begin{tabular}{ccl}
    \toprule
    Method&Score on test set\\
    \midrule
    SCIBERT & 0.38+\\
    +Redundancy removal & 0.408+\\
    +CS\_ROBERTA\_BASE & 0.43+\\
    +Title & 0.447+\\
    +Content prompt words & 0.452+\\
  \bottomrule
\end{tabular}
\end{table}

\subsection{GCN}
\quad Here we mainly introduce our experimental results on GCN. We tried GCN models with various architectures and different text embedding models.

\textbf{Details} We tried GCN models with different architectures, different numbers of layers, and different numbers of hidden units. We also tried GAT and RGCN based on heterogeneous relational graphs. In terms of experimental results, they seem to be not much different. We first divide the paper into semantic-based chunks with a chunk size of 300 characters. Then these text chunks are mapped to a high-dimensional semantic space as the node information of the GNN through an embedding model. We train our model on an RTX4090. epoch=100.

\textbf{Results} 
The experimental results are shown in Table 2. We found that the main factors affecting the effect of GCN are the way of graph construction and the choice of embedding model. The mapping dimension of the embedding model should not be greater than 1024, and the graph construction should not be too complex. The number of GCN layers should not be too many.

In addition to the information shown in Table 2. We also made other attempts on the size of the text chunk and the way of graph construction. The results show that the size of the text chunk is best between 200 and 300. We also tried to decompose the introduction into more chunks and build edges with more references, but the effect showed a downward trend.

\begin{table}
  \caption{Experimental results of different methods on BERT-like models}
  \label{tab:freq}
  \centering
  \begin{tabular}{cccc}
    \toprule
    Model & Number of Layers & Encoder & Score on test set\\
    \midrule
    GCN & 2 & BGE\_M3 & 0.447\\
    GCN & 5 & BGE\_small & 0.426\\
    GCN & 2 & BGE\_small & 0.438\\
    GAT & 1 & BGE\_small & 0.440\\
    RGCN & 4 & BGE\_small & 0.437\\
    \bottomrule
  \end{tabular}
\end{table}

\subsection{Ensemble}
\quad At the end, we integrated the results from different solutions. Compared with the results of the original single solution, the results were significantly improved, which reflects the advantages of ensemble learning. The integration results are shown in Table 3.

In the final stage, our solution mainly uses the integration of three solutions, including ROBERTA text classification based on semantic segmentation (Roberta-spacy), ROBERTA text classification based on absolute position segmentation (Roberta-absolute), and GCN-based text node classification.

\begin{table}
  \caption{The results of integrating different solutions in test stage}
  \label{tab:freq}
  \centering
  \begin{tabular}{ccl}
    \toprule
    Method & Score\\
    \midrule
    GCN + Roberta-spacy & 0.4783\\
    GCN + Roberta-spacy + Roberta-absolute & 0.4796\\
    \bottomrule
  \end{tabular}
\end{table}

\section{Other attempts}

\quad In addition to the above two solutions based on BERT and GCN, we also tried a solution based on a large language model~\cite{team2024chatglm} in the validation set stage. The effect of this solution was reduced in the test set stage, but due to the lack of hardware and time resources, we did not continue to try the large model in the test set stage. In the validation set stage, we input reference fragments and paper abstracts into the large language model, and by modifying the prompts, the large model can break through the original text limitations and output more in-depth text, and then input it into the BERT-like model for training. As shown in Figure 3. 

The results of the large model processing are shown in Table 4.

\begin{table}
  \caption{The results of the large model processing}
  \label{tab:freq}
  \centering
  \begin{tabular}{ccl}
    \toprule
    stage & Score\\
    \midrule
    Val & 0.47+\\
    Test & 0.40+\\
    \bottomrule
  \end{tabular}
\end{table}

\begin{figure}[h]
  \centering
  \includegraphics[width=\linewidth]{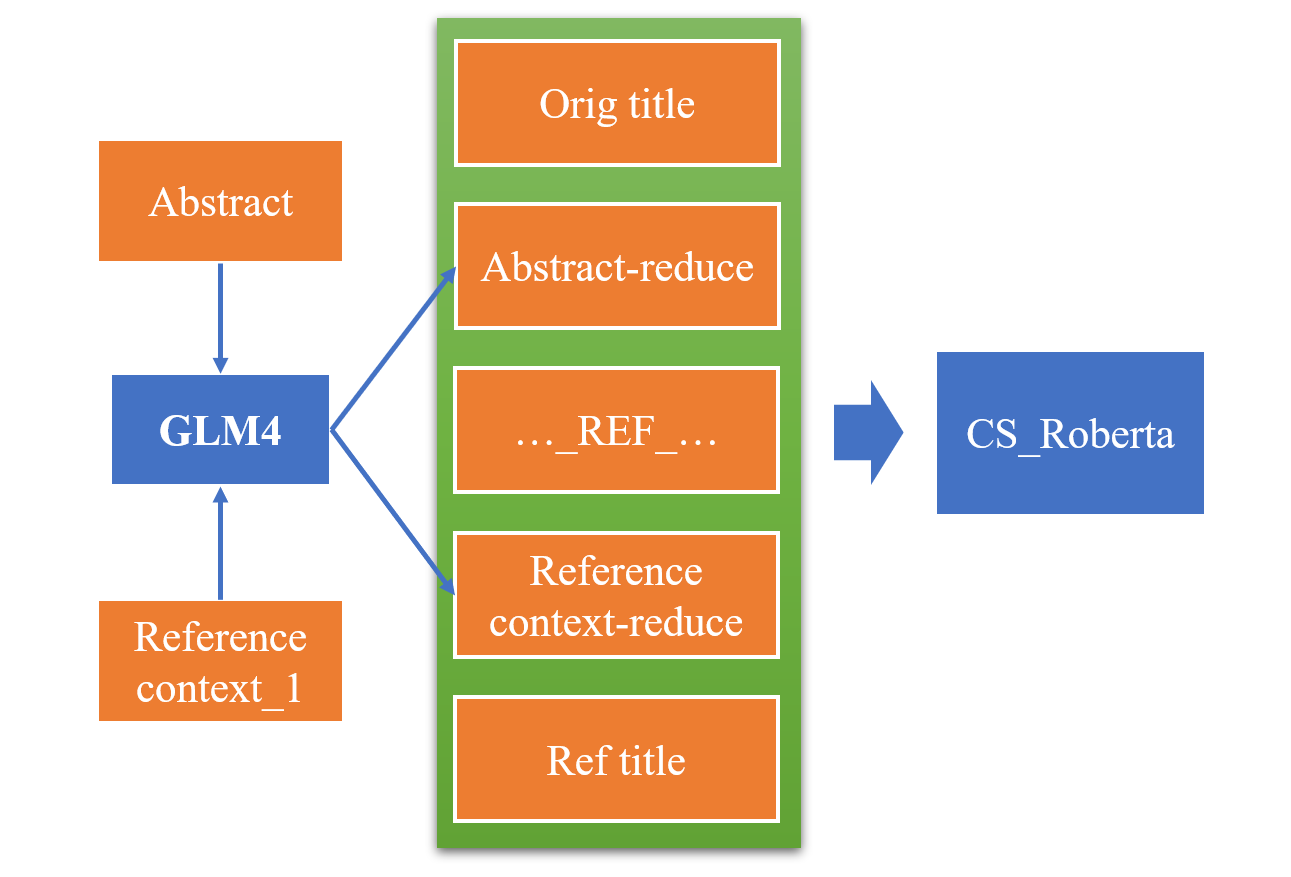}
  \caption{Training a BERT-like model using a large language model-processed dataset}
\end{figure}

\printbibliography


\end{document}